\documentclass[preprint,12pt]{elsarticle}

\usepackage[T1]{fontenc}
\usepackage{lmodern}

\usepackage{amsmath,amssymb}
\usepackage{graphicx}
\usepackage{textcomp}
\usepackage{array}
\usepackage{tabularx}
\usepackage{longtable}
\usepackage{algorithm}
\usepackage{algpseudocode}
\usepackage{orcidlink}
\usepackage{hyperref}
\hypersetup{hidelinks}

\biboptions{numbers,sort&compress}

\journal{arXiv}

\begin{document}

\begin{frontmatter}

\title{Reinforcement Learning and Rule-Based Peer-to-Peer Pricing in Residential PV-BES Communities}

\author[mineral]{Pablo Benalcazar\corref{cor1}\orcidlink{0000-0001-9578-8299}}
\ead{benalcazar@min-pan.krakow.pl}
\cortext[cor1]{Corresponding author.}

\author[mineral]{Maciej Kalka\orcidlink{0000-0002-4020-1582}}

\author[espoch]{Wilian Guam{\'a}n\orcidlink{0000-0002-9905-8231}}

\author[mineral]{Jacek Kami{\'n}ski\orcidlink{0000-0001-7514-8761}}

\affiliation[mineral]{organization={Division of Energy Economics, Mineral and Energy Economy Research Institute, Polish Academy of Sciences},
            city={Krak{\'o}w},
            country={Poland}}

\affiliation[espoch]{organization={GITEA, Escuela Superior Polit{\'e}cnica de Chimborazo (ESPOCH)},
            city={Riobamba},
            country={Ecuador}}

\begin{abstract}
This paper compares rule-based and learning-based pricing mechanisms for peer-to-peer (P2P) electricity trading in residential photovoltaic communities. 
The rule-based benchmarks comprise bill-sharing as an ex post allocation mechanism, the mid-market rate, and supply--demand-ratio pricing. 
The reinforcement-learning (RL) formulation is implemented through a Deep Q-Network and evaluated under multiplier-based and learnable SDR-shaped pricing, with a fixed-parameter SDR variant as a non-learning control. 
Performance is assessed through community savings together with complementary financial and operational indicators. 
In the base PV-only configuration, the rule-based benchmarks outperform the best RL policy. 
With battery energy storage, evaluated for the RL policies only, community savings under the best RL policy increase from \texteuro{}734.23 to \texteuro{}978.52. 
Across the learning-based modes and in both configurations, SDR-shaped pricing outperforms the multiplier-based parameterization considered. 
The results indicate that rule-based pricing remains highly competitive wherever the two families are compared directly, and that storage substantially improves the learning-based outcomes under this accounting, while the distribution of benefits remains heterogeneous across households.
\end{abstract}


\end{frontmatter}

\section*{Nomenclature}
\addcontentsline{toc}{section}{Nomenclature}
{\footnotesize
\setlength{\LTpre}{4pt}\setlength{\LTpost}{8pt}
\begin{longtable}{@{}p{0.17\textwidth}p{0.79\textwidth}@{}}
\hline
\multicolumn{2}{@{}l}{\textit{Abbreviations}} \\
BES & Battery energy storage \\
BS & Bill-sharing mechanism \\
DQN & Deep Q-Network \\
MMR & Mid-market rate \\
P2P & Peer-to-peer \\
PV & Photovoltaic \\
RL & Reinforcement learning \\
RL-M & RL-Multiplier: multiplier-based pricing \\
RL-SDR-F & RL-SDR-Fixed: fixed-parameter SDR-shaped pricing, non-learning control \\
RL-SDR-L & RL-SDR-Learnable: learnable SDR-shaped pricing \\
SDR & Supply--demand ratio \\
SOC & State of charge \\
SSI & Self-sufficiency index \\
\hline
\multicolumn{2}{@{}l}{\textit{Indices and sets}} \\
$t$ & Settlement interval (hourly) \\
$i$ & Household index; also the minibatch sample index in Algorithm~\ref{alg:dqn_training} \\
$k$ & Action index, $k=1,\ldots,K$ \\
$\mathcal{A}$ & Discrete action set of the RL agent \\
$K$ & Number of actions in $\mathcal{A}$ \\
$\mathcal{D}$ & Dataset of days with hourly PV, load, and grid-price profiles \\
$\mathcal{T}^{\mathrm{tr}}$ & Set of intervals with internal trading \\
$\mathcal{R}$ & Experience replay buffer \\
$\mathcal{B}$ & Minibatch sampled from $\mathcal{R}$ \\
\hline
\multicolumn{2}{@{}l}{\textit{Energy quantities} (kWh)} \\
$E_t^{\mathrm{sup}}$ & Aggregate local surplus available for internal trading \\
$E_t^{\mathrm{sup,pv}}$ & PV share of the aggregate local surplus \\
$E_t^{\mathrm{dem}}$ & Aggregate residual internal demand \\
$E_t^{\mathrm{tr}}$ & Internally traded volume, $E^{\mathrm{tr}}$ annual total \\
$E_t^{\mathrm{buy}}$ & Energy purchased internally by buyers \\
$E_t^{\mathrm{sell}}$ & Energy sold internally by sellers \\
$E_t^{\mathrm{pv}}$ & Community PV generation \\
$E_t^{\mathrm{load}}$ & Community household consumption \\
$E_t^{\mathrm{grid,in}}$ & Grid imports, $E^{\mathrm{grid,in}}$ annual total \\
$E_t^{\mathrm{grid,out}}$ & Grid exports \\
$b_t$ & Community net balance, $b_t=E_t^{\mathrm{pv}}-E_t^{\mathrm{load}}$ \\
$\mathrm{SDR}_t$ & Supply--demand ratio, $E_t^{\mathrm{sup}}/E_t^{\mathrm{dem}}$ (--) \\
\hline
\multicolumn{2}{@{}l}{\textit{Prices} (\texteuro{}/kWh)} \\
$p_t^{\mathrm{buy}}$ & Grid import price \\
$p_t^{\mathrm{sell}}$ & Grid export price \\
$p_t^{\mathrm{MMR}}$ & Mid-market rate internal price \\
$p_t^{\mathrm{SDR,sell}}$ & SDR benchmark internal sell price \\
$p_t^{\mathrm{SDR,buy}}$ & SDR benchmark average buy price \\
$p_t^{\mathrm{p2p,sell}}$ & Internal sell price quoted by the RL agent \\
$p_t^{\mathrm{p2p,buy}}$ & Internal buy price quoted by the RL agent \\
$p_t^{\mathrm{tr}}$ & Realized internal trading price, $\bar{p}^{\mathrm{tr}}$ its average \\
\hline
\multicolumn{2}{@{}l}{\textit{Costs and economic indicators} (\texteuro{})} \\
$C^{\mathrm{ref}}$ & Community cost without internal trading, $C_i^{\mathrm{ref}}$ per household \\
$C^{\mathrm{pool}}$ & Community cost after internal trading \\
$C_i^{\mathrm{BS}}$ & Cost assigned to household $i$ under bill-sharing \\
$B^{\mathrm{comm}}$ & Community benefit, $C^{\mathrm{ref}}-C^{\mathrm{pool}}$ (distinct from the minibatch size $B$) \\
$\omega_i$ & Share of the community benefit allocated to household $i$, $\sum_i\omega_i=1$ (--) \\
$C^{\mathrm{usr}}$ & User cost \\
$R^{\mathrm{pro}}$ & Prosumer revenue \\
$S^{\mathrm{comm}}$ & Community savings \\
$J_t^{\mathrm{base}}$ & Hourly community cost of the no-P2P reference \\
$J_t^{\mathrm{p2p}}$ & Hourly community cost after P2P clearing and grid settlement \\
\hline
\multicolumn{2}{@{}l}{\textit{Pricing-mechanism parameters}} \\
$\mu^{\mathrm{buy}}_k,\ \mu^{\mathrm{sell}}_k$ & Buy and sell multipliers of action $k$ under RL-M (--) \\
$\alpha,\ \beta$ & Sell- and buy-price sensitivity of the SDR-shaped rule, $0\leq\alpha\leq1$, $\beta\geq1$ (--) \\
$C_{\mathrm{ctrl}}$ & Half-width of the grid price corridor (\texteuro{}/kWh) \\
$C_{\mathrm{bal}}$ & Midpoint of the grid price corridor (\texteuro{}/kWh) \\
$\delta$ & Log-argument stability constant (--) \\
\hline
\multicolumn{2}{@{}l}{\textit{Reinforcement-learning parameters}} \\
$s_t$ & State vector at interval $t$ \\
$\tilde{b}_t$ & Discretized community net balance, nine bins (--) \\
$\overline{\mathrm{SOC}}_t$ & Mean battery state of charge across households, as a fraction of capacity (--) \\
$a_t$ & Action selected at interval $t$ \\
$r_t$ & Reward, $J_t^{\mathrm{base}}-J_t^{\mathrm{p2p}}$ (\texteuro{}) \\
$h_t$ & Hour of the day \\
$Q_\theta$ & Action-value network with parameters $\theta$ \\
$Q_{\hat\theta}$ & Target network with parameters $\hat\theta$ \\
$\pi^*$ & Greedy policy obtained after training \\
$\gamma$ & Discount factor (--) \\
$\eta$ & Adam learning rate (--) \\
$B$ & Minibatch size (--) \\
$T$ & Target-network synchronization period (steps) \\
$E$ & Number of training episodes (--) (distinct from the energy quantities $E_t^{\bullet}$) \\
$\varepsilon$ & Exploration rate, from $\varepsilon_0$ to $\varepsilon_{\min}$ (--) \\
$\lambda$ & Exploration decay rate per step (--) \\
\hline
\end{longtable}}

\section{Introduction}\label{sec:introduction}

Peer-to-peer (P2P) electricity trading allows prosumers and consumers to exchange electricity locally instead of relying exclusively on the external grid~\cite{Sousa2019}. It forms part of the broader development of local energy markets, where participation rules, governance structures, and settlement mechanisms continue to evolve~\cite{Capper2022}. Within that setting, pricing becomes particularly important because it shapes both the value of internal transactions and the way economic gains are shared among participants.

Among the pricing mechanisms proposed for P2P electricity trading, the supply--demand ratio (SDR) and the mid-market rate (MMR) are the most widely discussed~\cite{Kim2023}. Their selection is not neutral, as previous studies show that different pricing rules can materially affect consumer payments and prosumer revenues~\cite{Zhou2018}. In residential settings, those effects also depend on available technologies, tariff design, and solar generation conditions~\cite{Neves2020,Benalcazar2024}.

Reinforcement learning (RL) is increasingly being used in power-system operation and planning~\cite{Pesantez2024}. In P2P electricity markets, RL has been applied to price adjustment and trading coordination under changing operating conditions~\cite{Qiu2021}. This is particularly relevant, as learning-based policies may improve aggregate economic outcomes. However, their advantage over simpler rule-based benchmarks remains unclear~\cite{May2023}.

In local communities with photovoltaic (PV) generation, the economic performance of P2P trading depends on the local surplus available for exchange and on the technical configuration of the community, since changes in PV output affect self-sufficiency, economic savings, and trading revenues~\cite{Neves2020,Huang2022}. In this context, battery storage plays a key role by influencing both the internal exchange conditions and the resulting economic outcomes.

Rule-based mechanisms and learning-based schemes have both been explored in the literature~\cite{Capper2022,May2023}, yet comparative evidence focused on their financial outcomes remains limited~\cite{Islam2024}. This paper examines rule-based and learning-based pricing formulations in a P2P electricity market. The rule-based benchmarks are bill-sharing, MMR, and SDR, while the RL formulation is evaluated through three operating modes. The rule-based mechanisms are assessed in the base PV-only framework, whereas the RL policies are evaluated in both PV-only and PV-BES configurations. The comparison is based on average trading price, user cost, prosumer revenue, community savings, and complementary indicators of internal trading and self-sufficiency.

\section{Market Framework and Pricing Mechanisms}\label{sec:framework}

\subsection{P2P Market Framework}

A local electricity community is considered to comprise prosumers with PV generation and consumers without on-site generation \cite{Capper2022,Sousa2019}. At each interval, the energy exchanged internally is limited by the balance between aggregate local surplus and aggregate internal demand.

Let $E_t^{\mathrm{sup}}$ denote the total surplus available for local trading and let $E_t^{\mathrm{dem}}$ denote the total residual internal demand at interval $t$, both net of self-consumption. The traded volume is defined as
\begin{equation}
E_t^{\mathrm{tr}} = \min \left(E_t^{\mathrm{sup}}, E_t^{\mathrm{dem}}\right).
\label{eq:traded}
\end{equation}

Equation~\eqref{eq:traded} gives the internally matchable volume. The rule-based benchmarks settle this volume in full in every interval, with internal prices bounded by the grid import and export prices, which ensures that neither side is worse off than when settling with the grid whenever the export price lies below the import price. The RL-based mechanisms instead quote prices ex ante, and internal exchange is settled only in intervals where the quoted prices remain strictly inside the grid price corridor and the internal buy--sell spread is non-negative. Otherwise no internal exchange takes place in that interval and both sides settle directly with the grid. In the intervals where the export price exceeds the import price, full rule-based settlement reduces the community benefit, whereas the RL-based mechanisms do not settle those intervals internally. The realized annual volume therefore depends on the pricing mechanism, even though~\eqref{eq:traded} itself does not. The residual demand is covered by grid imports, while surplus not absorbed internally is exported to the grid. Within each interval, the traded volume is allocated to sellers in proportion to their surplus and to buyers in proportion to their residual demand.

\subsection{Benchmark Pricing and Settlement Mechanisms}

\subsubsection{Bill-Sharing Mechanism}
Bill-sharing represents collective settlement through the redistribution of the community benefit \cite{Gonzalez-Asenjo2023}. Let $C^{\mathrm{ref}}$ denote the total community cost without internal trading, and let $C^{\mathrm{pool}}$ denote the total cost after local trading. The community benefit is
\begin{equation}
B^{\mathrm{comm}} = C^{\mathrm{ref}} - C^{\mathrm{pool}}.
\end{equation}
The cost assigned to participant $i$ is then
\begin{equation}
C_i^{\mathrm{BS}} = C_i^{\mathrm{ref}} - \omega_i B^{\mathrm{comm}},
\qquad \sum_i \omega_i = 1,
\end{equation}
where $\omega_i$ denotes the allocated share of the community benefit, set in proportion to each household's share of the internally matched energy.

\subsubsection{Mid-Market Rate Pricing}
The mid-market rate mechanism defines an internal price between the grid import and export prices \cite{Tushar2018}. Let $p_t^{\mathrm{buy}}$ and $p_t^{\mathrm{sell}}$ denote the grid import and export prices at interval $t$. The MMR price is
\begin{equation}
p_t^{\mathrm{MMR}} = \frac{p_t^{\mathrm{buy}} + p_t^{\mathrm{sell}}}{2}.
\end{equation}
This benchmark maintains a direct relation to the external grid price corridor.

\subsubsection{Supply--Demand Ratio Pricing}
In the SDR rule, the internal price depends on the local market balance \cite{Liu2017,Zhou2018}. The supply--demand ratio at interval $t$ is
\begin{equation}
\mathrm{SDR}_t = \frac{E_t^{\mathrm{sup}}}{E_t^{\mathrm{dem}}}.
\label{eq:sdr}
\end{equation}
The corresponding internal sell and buy prices follow the uniform pricing model of \cite{Zhang2023}, in which both prices decrease monotonically from the grid buy price when local supply is scarce to the grid sell price when supply meets demand, giving
\begin{equation}
p_t^{\mathrm{SDR,sell}} =
\begin{cases}
\dfrac{p_t^{\mathrm{buy}}\,p_t^{\mathrm{sell}}}{\left(p_t^{\mathrm{buy}} - p_t^{\mathrm{sell}}\right)\mathrm{SDR}_t + p_t^{\mathrm{sell}}}, & 0 \leq \mathrm{SDR}_t \leq 1, \\[4pt]
p_t^{\mathrm{sell}}, & \mathrm{SDR}_t > 1,
\end{cases}
\label{eq:sdr_sell}
\end{equation}
\begin{equation}
p_t^{\mathrm{SDR,buy}} =
\begin{cases}
p_t^{\mathrm{SDR,sell}}\,\mathrm{SDR}_t + p_t^{\mathrm{buy}}\left(1-\mathrm{SDR}_t\right), & 0 \leq \mathrm{SDR}_t \leq 1, \\
p_t^{\mathrm{sell}}, & \mathrm{SDR}_t > 1,
\end{cases}
\label{eq:sdr_buy}
\end{equation}
so that the price responds to local scarcity and surplus conditions within the grid price bounds. Internal exchange executes at the single price $p_t^{\mathrm{SDR,sell}}$, and each buyer covers its residual shortage at the grid import price. The buy price in~\eqref{eq:sdr_buy} is therefore the average a buyer pays across its internal and residual purchases rather than a separately executed price, which makes the mechanism budget-balanced and leaves no spread retained outside the community. This is the settlement of \cite{Zhang2023} and of the original supply--demand-ratio construction of \cite{Liu2017}.

\subsection{RL-Based Pricing}
Unlike the rule-based benchmarks, the RL formulation updates internal prices sequentially, with a centralized agent that observes the community state at each settlement interval and selects the pricing parameters. The study evaluates three parameterizations of this mapping; in two of them the agent selects among candidate actions, while the third holds the parameters fixed and serves as a non-learning control within the same pricing family.

\subsubsection{Multiplier-Based Pricing} 
Each action maps to a pair of scaling factors $(\mu^{\mathrm{buy}}_k, \mu^{\mathrm{sell}}_k)$ applied directly to the prevailing grid prices, so that
\begin{align}
p_t^{\mathrm{p2p,buy}}  &= \mu^{\mathrm{buy}}_k \cdot p_t^{\mathrm{buy}}, \\
p_t^{\mathrm{p2p,sell}} &= \mu^{\mathrm{sell}}_k \cdot p_t^{\mathrm{sell}}.
\end{align}
This mode is denoted RL-Multiplier (RL-M), and the candidate values are listed in Table~\ref{tab:pricing_params}, paired index-wise to give $K=8$ actions. The settlement condition introduced in Sect.~\ref{sec:framework} requires $p_t^{\mathrm{sell}} < p_t^{\mathrm{p2p,sell}} \leq p_t^{\mathrm{p2p,buy}} < p_t^{\mathrm{buy}}$. Because the two multipliers scale different grid prices, RL-M does not guarantee the middle inequality. Whenever $\mu^{\mathrm{buy}}_k\,p_t^{\mathrm{buy}} < \mu^{\mathrm{sell}}_k\,p_t^{\mathrm{sell}}$, the internal spread inverts, the interval is not settled internally, and both sides trade with the grid.

\subsubsection{SDR-Based Pricing} Building on the supply--demand ratio defined in~\eqref{eq:sdr}, the P2P prices are computed as
\begin{align}
p_t^{\mathrm{p2p,sell}} &= C_{\mathrm{ctrl}} \cdot \tanh\!\bigl(-\alpha\ln(\delta{+}\mathrm{SDR}_{t-1})\bigr) + C_{\mathrm{bal}}, \label{eq:p2p_sell} \\
p_t^{\mathrm{p2p,buy}}  &= C_{\mathrm{ctrl}} \cdot \tanh\!\bigl(-\beta\ln(\delta{+}\mathrm{SDR}_{t-1})\bigr) + C_{\mathrm{bal}}, \label{eq:p2p_buy}
\end{align}
where $C_{\mathrm{ctrl}} = (p_t^{\mathrm{buy}}{-}p_t^{\mathrm{sell}})/2$, $C_{\mathrm{bal}} = (p_t^{\mathrm{buy}}{+}p_t^{\mathrm{sell}})/2$, and $\delta > 0$ is a small stability constant. 
Moreover, if $\mathrm{SDR}_{t-1} > 1$, the price in~\eqref{eq:p2p_buy} is set to $C_{\mathrm{bal}}$ to preserve a non-negative spread for sellers when local supply exceeds demand.
Prices for interval $t$ are posted ex ante from the community state observed at the start of the interval. The supply--demand ratio entering~\eqref{eq:p2p_sell} and~\eqref{eq:p2p_buy} is therefore the ratio realized in interval $t-1$, and the same convention applies to the state vector defined below. At the start of each episode, the lagged supply--demand ratio is initialized to one.

The prices remain bounded between $p_t^{\mathrm{sell}}$ and $p_t^{\mathrm{buy}}$, with the bounds reached under extreme scarcity or surplus, which preserves individual rationality for both buyers and sellers. The pair $(\alpha,\beta)$, where $0 \leq \alpha \leq 1$ and $\beta \geq 1$, determines how sensitive the pricing rule is to changes in the supply--demand balance. 
Under~RL-SDR-Fixed~(RL-SDR-F), the same pair $(\alpha,\beta)$ is used throughout the horizon. Its action set contains a single element. Consequently, no learning takes place and the state cannot influence the resulting price. RL-SDR-F therefore acts as a fixed-parameter control that isolates the contribution of adapting $\alpha$ to operating conditions. Under RL-SDR-Learnable (RL-SDR-L), the agent chooses $(\alpha,\beta)$ from the discrete set $\mathcal{A}=\{(\alpha_k,\beta_k)\}_{k=1}^{K}$, allowing the response to change with operating conditions. In both cases, the supply--demand ratio remains the underlying signal, but the resulting prices do not coincide with the rule-based SDR benchmark because the mapping is different. Table~\ref{tab:pricing_params} reports the value of $\delta$ and the candidate sensitivity pairs.
All three pricing modes share the same state description, although in RL-SDR-F the state cannot affect the selected action. The community net balance between the PV generation $E_t^{\mathrm{pv}}$ and the household consumption $E_t^{\mathrm{load}}$ at interval $t$ is
\begin{equation}
b_t = E_t^{\mathrm{pv}} - E_t^{\mathrm{load}},
\end{equation}
and its discretized version, $\tilde{b}_t$, with nine bins whose edges lie at $\pm0.2$, $\pm1$, $\pm2$, and $\pm5$~kWh, is used to keep the state compact. The mean battery state of charge $\overline{\mathrm{SOC}}_t$ is taken over all households as a fraction of capacity, with households without storage contributing zero, and is discretized into five equal bins, while the supply--demand ratio enters through a bounded logarithmic transformation. Together with the cyclic encoding of the hour, the state is
\begin{equation}
s_t = \bigl(\sin(2\pi h_t/24),\;\cos(2\pi h_t/24),\;\tilde{b}_t,\;\overline{\mathrm{SOC}}_t,\;\mathrm{SDR}_{t-1}\bigr).
\end{equation}
For PV-only, $\overline{\mathrm{SOC}}_t$ is set to zero, which keeps the same state structure in both technical configurations.
PV generation does not appear as a separate feature because its effect is already captured by $b_t$ and $\mathrm{SDR}_{t-1}$, both of which reflect the surplus available for local exchange. Because load, generation, and the battery trajectory are exogenous, the pricing action does not affect the state transition, and the DQN thus serves as a discrete-action solver for the learning problem. In every mode, the reward corresponds to the hourly cost saving relative to a no-P2P reference, defined as
\begin{equation}
r_t = J_t^{\mathrm{base}} - J_t^{\mathrm{p2p}},
\end{equation}
where $J_t^{\mathrm{base}} = E_t^{\mathrm{dem}}\,p_t^{\mathrm{buy}} - E_t^{\mathrm{sup,pv}}\,p_t^{\mathrm{sell}}$, with $E_t^{\mathrm{sup,pv}}$ the PV share of the surplus, and $J_t^{\mathrm{p2p}}$ is the actual community cost after P2P clearing and residual grid settlement. The RL agent is therefore trained to maximize aggregate community savings. Because the SDR-shaped and multiplier-based rules quote separate buy and sell prices, an interval with $p_t^{\mathrm{p2p,buy}} > p_t^{\mathrm{p2p,sell}}$ generates a settlement spread that is retained by the market operator and is not redistributed to participants. This spread enters $J_t^{\mathrm{p2p}}$ as a cost. As a result, maximizing community savings implicitly rewards a narrower internal spread as well as a larger internal exchange. In the PV-BES configuration, battery energy offered for internal exchange has no grid-export alternative. For this reason, $J_t^{\mathrm{base}}$ credits export revenue only for the PV surplus. In intervals without internal settlement, the battery energy offered for exchange is curtailed, while the PV surplus is exported to the grid. The training baseline therefore includes battery operation, whereas the reported savings are expressed relative to $C^{\mathrm{ref}}$, which excludes the battery.

The learning stage relies on a Deep Q-Network (DQN)~\cite{Mnih2015} to approximate the action-value function $Q_\theta(s,a)$ through a feedforward neural network. To stabilize training, the model uses a target network $Q_{\hat\theta}$ and an experience replay buffer $\mathcal{R}$, from which mini-batches are drawn to minimize the mean-squared Bellman error. 
The network uses two hidden layers of 128 and 64 rectified-linear units and a
linear output of $|\mathcal{A}|$ units, trained with Adam ($\eta{=}5{\times}10^{-4}$) on minibatches of $B{=}64$ from a replay buffer of capacity 50,000.
The discount factor is $\gamma{=}0.95$, and the target network is synchronized every $T{=}200$ steps.
Transitions are bootstrapped uniformly, including at the episode boundary. Because the transition process is exogenous, the bootstrapped term does not depend on the selected action and does not alter the greedy policy.
Training runs for $E{=}500$ episodes, each covering one day drawn uniformly at random, with replacement, from the 366 days of the annual dataset.
Exploration follows an $\varepsilon$-greedy schedule
decaying exponentially from $1.0$ to $0.01$ at rate $\lambda{=}0.99952$ per step.
Both scenarios share these hyperparameters and differ only in their technical configuration.
The full training procedure is outlined in Algorithm~\ref{alg:dqn_training}.
Policies are evaluated greedily over the same annual dataset used for training. The reported results come from a single training run per mode and configuration, with a fixed random seed. The differences among the RL modes therefore reflect a single realization of the training process.

\begin{table}[t]
\caption{Pricing parameters for each pricing mode.}\label{tab:pricing_params}
\centering
\footnotesize
\renewcommand{\arraystretch}{1.0}
\setlength{\tabcolsep}{4pt}
\begin{tabularx}{\textwidth}{@{}p{0.31\textwidth}p{0.16\textwidth}X@{}}
\hline
\textbf{Parameter} & \textbf{Symbol} & \textbf{Value} \\
\hline
\multicolumn{3}{@{}l}{\textit{SDR stability constant (RL-SDR-F and RL-SDR-L)}} \\
Log-argument guard & $\delta$ & $10^{-6}$ \\
\hline
\multicolumn{3}{@{}l}{\textit{RL-M action set ($K=8$)}} \\
Buy multipliers & $\mu^{\mathrm{buy}}_k$ & $0.2,\ 0.3,\ \ldots,\ 0.9$ \\
Sell multipliers & $\mu^{\mathrm{sell}}_k$ & $1.05,\ 1.10,\ \ldots,\ 1.40$ \\
\hline
\multicolumn{3}{@{}l}{\textit{RL-SDR-F ($K=1$)}} \\
Fixed sensitivity & $(\alpha,\beta)$ & $(0.5,\ 2.0)$ \\
\hline
\multicolumn{3}{@{}l}{\textit{RL-SDR-L action set ($K=9$)}} \\
Sensitivity pairs & $(\alpha_k,\beta_k)$ & $(0.1,\ 2.0),\ldots,(0.9,\ 2.0)$ \\
\hline
\end{tabularx}
\end{table}

\begin{algorithm}[t]
\caption{DQN training for RL-based P2P pricing.}\label{alg:dqn_training}
\footnotesize
\begin{algorithmic}[1]
\Require Dataset $\mathcal{D}$ of days with hourly PV, load, and grid-price profiles; action set $\mathcal{A}$
\Require Discount $\gamma$; batch size $B$; target-update period $T$; exploration rate $\varepsilon_0$; decay $\lambda$; minimum $\varepsilon_{\min}$
\State Initialize $Q_\theta$ randomly; set $Q_{\hat\theta}\gets Q_\theta$, $\mathcal{R}\gets\emptyset$, $n\gets0$, and $\varepsilon\gets\varepsilon_0$
\For{episode $e=1,\ldots,E$}
    \State Sample day $d\sim\mathcal{D}$ uniformly with replacement, load its 24 hourly profiles, reset the community, and observe $s_0$
    \For{$t=0,\ldots,23$}
        \If{$\mathrm{Uniform}(0,1)<\varepsilon$}
            \State $a_t\gets\mathrm{random}(\mathcal{A})$ \Comment{explore}
        \Else
            \State $a_t\gets\arg\max_a Q_\theta(s_t,a)$ \Comment{exploit}
        \EndIf
        \State Compute $(p_t^{\mathrm{p2p,buy}},p_t^{\mathrm{p2p,sell}})$ for the active pricing mode
        \State Step the environment and observe $s_{t+1}$, $J_t^{\mathrm{p2p}}$, and $J_t^{\mathrm{base}}$
        \State $r_t\gets J_t^{\mathrm{base}}-J_t^{\mathrm{p2p}}$ and store $(s_t,a_t,r_t,s_{t+1})$ in $\mathcal{R}$
        \If{$|\mathcal{R}|\geq B$}
            \State Sample minibatch $\mathcal{B}=\{(s_i,a_i,r_i,s'_i)\}\sim\mathcal{R}$
            \State $y_i\gets r_i+\gamma\max_{a'}Q_{\hat\theta}(s'_i,a')$ for all $i\in\mathcal{B}$
            \State $\theta\gets\mathrm{Adam}\!\left(\theta,\nabla_\theta\frac{1}{B}\sum_{i\in\mathcal{B}}\left(y_i-Q_\theta(s_i,a_i)\right)^2;\eta\right)$
        \EndIf
        \State $n\gets n+1$
        \If{$n\bmod T=0$}
            \State $\hat\theta\gets\theta$ \Comment{synchronize target network}
        \EndIf
        \State $\varepsilon\gets\max(\varepsilon_{\min},\varepsilon\lambda)$
    \EndFor
\EndFor
\Ensure Greedy policy $\pi^*(s)=\arg\max_a Q_\theta(s,a)$
\end{algorithmic}
\end{algorithm}

\subsection{Evaluation Metrics}

The comparison between mechanisms is carried out through financial and operational metrics. Because the RL reward is defined from aggregate community savings, community savings are considered the main economic outcome. Average trading price, user cost, prosumer revenue, traded energy, self-sufficiency, and grid imports are then used to interpret how each mechanism produces that outcome~\cite{Zhou2018,May2023}. Table~\ref{tab:metrics} lists the metrics used in the comparative analysis, in which $E_t^{\mathrm{buy}}$ and $E_t^{\mathrm{sell}}$ denote the energy bought and sold internally, $E_t^{\mathrm{grid,in}}$ and $E_t^{\mathrm{grid,out}}$ the community grid imports and exports, and $p_t^{\mathrm{tr}}$ the internal settlement price. For the rule-based benchmarks this is a single price that enters both the user cost and the prosumer revenue. The RL modes apply the internal buy price in the user cost and the internal sell price in the prosumer revenue, and $p_t^{\mathrm{tr}}$ then denotes their midpoint, which is used only in the average trading price over the set $\mathcal{T}^{\mathrm{tr}}$ of intervals with non-zero internal exchange. The self-sufficiency index uses the total household consumption $E_t^{\mathrm{load}}$, which includes the share met by on-site generation and differs from the residual demand $E_t^{\mathrm{dem}}$. The community savings coincide with the community benefit $B^{\mathrm{comm}}$.

\begin{table}[t]
\caption{Evaluation metrics considered in this study.}\label{tab:metrics}
\centering
\footnotesize
\setlength{\tabcolsep}{3pt}
\renewcommand{\arraystretch}{1.0}
\begin{tabularx}{\textwidth}{@{}p{0.24\textwidth}X>{\raggedleft\arraybackslash}p{0.06\textwidth}@{}}
\hline
\textbf{Metric} & \textbf{Equation} & \textbf{Ref.} \\
\hline
\shortstack[l]{Average trading\\price} & $\bar{p}^{\mathrm{tr}}=|\mathcal{T}^{\mathrm{tr}}|^{-1}\sum_{t\in\mathcal{T}^{\mathrm{tr}}}p_t^{\mathrm{tr}}$ & \cite{Zhou2018} \\
User cost & $C^{\mathrm{usr}}=\sum_t\left(p_t^{\mathrm{tr}}E_t^{\mathrm{buy}}+p_t^{\mathrm{buy}}E_t^{\mathrm{grid,in}}\right)$ & \cite{Zhou2018} \\
Prosumer revenue & $R^{\mathrm{pro}}=\sum_t\left(p_t^{\mathrm{tr}}E_t^{\mathrm{sell}}+p_t^{\mathrm{sell}}E_t^{\mathrm{grid,out}}\right)$ & \cite{Huang2022} \\
\shortstack[l]{Community\\savings} & $S^{\mathrm{comm}}=C^{\mathrm{ref}}-C^{\mathrm{pool}}$ & \cite{May2023} \\
\shortstack[l]{Self-sufficiency\\index} & $\mathrm{SSI}=1-\frac{\sum_tE_t^{\mathrm{grid,in}}}{\sum_tE_t^{\mathrm{load}}}$ & \cite{Huang2022} \\
Traded energy & $E^{\mathrm{tr}}=\sum_{t\in\mathcal{T}^{\mathrm{tr}}}E_t^{\mathrm{tr}}$ & \cite{Hoque2024} \\
Grid imports & $E^{\mathrm{grid,in}}=\sum_tE_t^{\mathrm{grid,in}}$ & \cite{Neves2020} \\
\hline
\end{tabularx}
\end{table}

\section{Case Study}\label{sec:case}

The case study considers a residential electricity community composed of 20 heterogeneous households. Local trading is evaluated over an annual horizon with an hourly settlement interval, and the remaining imbalance is settled against time-varying grid buy and sell prices, with all monetary values expressed in euro. Eight households are prosumers with on-site PV generation, while the remaining 12 households are consumers without local generation. The household demand profiles were synthetically generated using the RAMP framework, a bottom-up open-source stochastic model for load-profile generation~\cite{Lombardi2019}. The model was calibrated to represent Polish households using official energy statistics published by the Central Statistical Office of Poland~\cite{StatisticsPoland2026}. The hourly PV generation profiles are derived from an hourly irradiance and ambient-temperature series for Poland obtained from the Renewables.ninja platform~\cite{Pfenninger2016}, converted through a capacity-based model with temperature derating.

The import price follows a two-zone time-of-use tariff of the kind offered to Polish households under the G12w tariff group, equal to 0.249~\texteuro{}/kWh in the weekday peak zones from 06:00 to 13:00 and from 15:00 to 22:00, and to 0.137~\texteuro{}/kWh in all other hours, including weekends. The export price follows the hourly market price applied to prosumer settlement under the Polish net-billing scheme, ranging from zero to 0.657~\texteuro{}/kWh with an annual mean of 0.100~\texteuro{}/kWh. In 384 of the 8,784 hourly intervals the export price exceeds the import price, in which case the grid price corridor is inverted.

Two technical configurations are considered. The first corresponds to a PV-only community, while the second retains the same PV portfolio and adds battery storage to the eight prosumer households. Table~\ref{tab:case_study} summarizes the main characteristics of both configurations. Both cases consider the same community composition, settlement interval, and PV portfolio. The PV-BES configuration differs from the base PV-only case by including battery storage, with an aggregate energy capacity of 68.15~kWh and aggregate charge/discharge power of 32.07~kW. Battery sizing and hourly dispatch are determined upstream by a genetic algorithm that jointly optimizes each household's PV and battery capacity together with a rule-based dispatch policy, minimizing annualized system cost (capital recovery, operation and maintenance, grid trading, and battery degradation) over one representative year. The resulting hourly state-of-charge trace, sized assuming a round-trip efficiency of approx.\ 91.4\%, is a fixed input to the P2P evaluation and is not altered by the settlement mechanism or the RL policy. That dispatch is degradation-aware and cycles the batteries conservatively, with the result that the storage contribution reported below is bounded by the upstream sizing rather than by the settlement mechanism. Battery discharge enters the hourly community balance, whereas the corresponding charging energy is committed in the upstream dispatch and is not debited from the simulated PV surplus or imports. The reported PV-BES results therefore include the value of this upstream-committed charging energy. Both configurations are evaluated on the same realization of the household demand profiles, which keeps the comparison controlled and gives the two cases a common community reference cost $C^{\mathrm{ref}}$ of \texteuro{}4,747.81.

\begin{table}[t]
\caption{Case-study characteristics and technical configurations.}\label{tab:case_study}
\centering
\footnotesize
\renewcommand{\arraystretch}{1.0}
\setlength{\tabcolsep}{4pt}
\begin{tabularx}{\textwidth}{@{}Xcc@{}}
\hline
\textbf{Parameter} & \textbf{PV-only} & \textbf{PV-BES} \\
\hline
Households (prosumers/consumers) & 20 (8/12) & 20 (8/12) \\
Households with battery storage & 0 & 8 \\
Community installed PV capacity (kW) & 20.14 & 20.14 \\
Community BESS energy capacity (kWh) & 0.00 & 68.15 \\
Community BESS charge/discharge power (kW) & 0.00 & 32.07 \\
Total annual household consumption (kWh) & 37,742.06 & 37,742.06 \\
Mean annual consumption per household (kWh) & 1,887.1 & 1,887.1 \\
Mean annual consumption of prosumers (kWh/household) & 1,960.4 & 1,960.4 \\
Mean annual consumption of consumers (kWh/household) & 1,838.2 & 1,838.2 \\
\hline
\end{tabularx}
\end{table}

\section{Results and Discussion}\label{sec:results}
\subsection{Community-Level Comparison}

\begin{table}[t]
\caption{Financial community-level results.}\label{tab:results_financial}
\centering
\scriptsize
\renewcommand{\arraystretch}{1.0}
\setlength{\tabcolsep}{2.5pt}
\begin{tabular}{@{}llrrrrr@{}}
\hline
\textbf{Method} & \textbf{Case} & \shortstack{\textbf{Avg. price}\\\textbf{(\texteuro{}/kWh)}} & \shortstack{\textbf{User cost}\\\textbf{(\texteuro{})}} & \shortstack{\textbf{Pros. rev.}\\\textbf{(\texteuro{})}} & \shortstack{\textbf{Savings}\\\textbf{(\texteuro{})}} & \shortstack{\textbf{Savings}\\\textbf{(\%)}} \\
\hline
BS       & PV-only & ---   & ---      & ---      & 829.98   & 17.48 \\
MMR      & PV-only & 0.141 & 6,043.68 & 2,125.85 & 829.98   & 17.48 \\
SDR      & PV-only & 0.106 & 5,673.79 & 1,755.96 & 829.98   & 17.48 \\
RL-M     & PV-only & 0.120 & 6,126.61 & 1,797.29 & 418.49   & 8.81 \\
RL-SDR-F & PV-only & 0.140 & 6,095.92 & 1,964.36 & 616.26   & 12.98 \\
RL-SDR-L & PV-only & 0.147 & 6,095.92 & 2,082.34 & 734.23   & 15.46 \\
RL-M     & PV-BES  & 0.135 & 6,018.64 & 1,877.76 & 606.94   & 12.78 \\
RL-SDR-F & PV-BES  & 0.150 & 5,981.40 & 2,093.68 & 860.09   & 18.12 \\
RL-SDR-L & PV-BES  & 0.156 & 5,981.40 & 2,212.11 & 978.52   & 20.61 \\
\hline
\end{tabular}
\end{table}

\begin{table}[t]
\caption{Operational community-level results.}\label{tab:results_operational}
\centering
\footnotesize
\renewcommand{\arraystretch}{1.0}
\setlength{\tabcolsep}{5pt}
\begin{tabular}{@{}llrrr@{}}
\hline
\textbf{Method} & \textbf{Case} & \textbf{SSI} & \shortstack{\textbf{Traded energy}\\\textbf{(kWh)}} & \shortstack{\textbf{Grid imports}\\\textbf{(kWh)}} \\
\hline
BS       & PV-only & 0.304 & 7,266 & 26,263 \\
MMR      & PV-only & 0.304 & 7,266 & 26,263 \\
SDR      & PV-only & 0.304 & 7,266 & 26,263 \\
RL-M     & PV-only & 0.219 & 4,069 & 29,459 \\
RL-SDR-F & PV-only & 0.301 & 7,131 & 26,398 \\
RL-SDR-L & PV-only & 0.301 & 7,131 & 26,398 \\
RL-M     & PV-BES  & 0.245 & 4,624 & 28,479 \\
RL-SDR-F & PV-BES  & 0.325 & 7,644 & 25,459 \\
RL-SDR-L & PV-BES  & 0.325 & 7,644 & 25,459 \\
\hline
\end{tabular}
\end{table}

\begin{figure}[t]
    \centering
    \includegraphics[width=\textwidth]{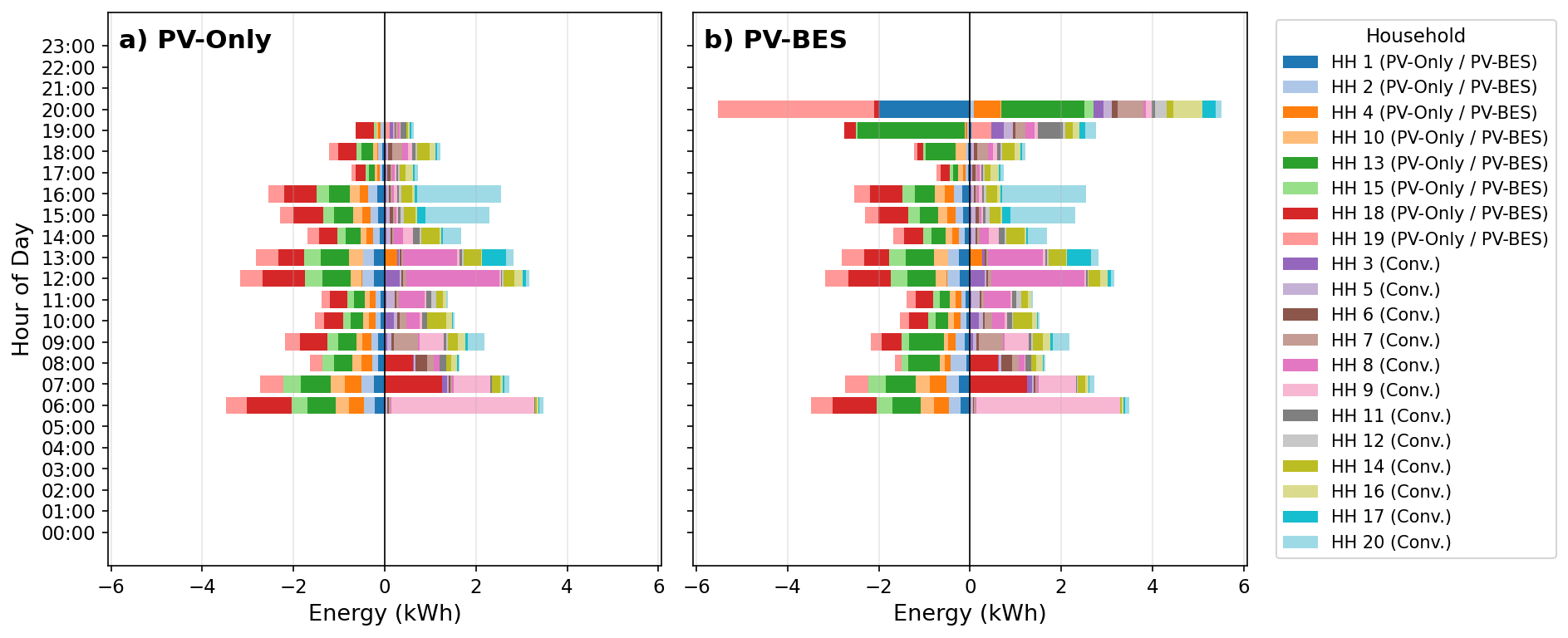}
    \caption{Hourly household-level P2P participation under RL-SDR-L on June 18 in the PV-only configuration (a) and the PV-BES configuration (b).}
    \label{fig:participation_sdr_l_pvbes}
\end{figure}

Table~\ref{tab:results_financial} reports the financial results, with percentage savings expressed relative to the community reference cost $C^{\mathrm{ref}}$. The BS row reports only savings because bill-sharing settles ex post without an hourly internal price. At the community level, the rule-based benchmarks perform best in the base PV-only configuration, with BS, MMR, and SDR all attaining \texteuro{}829.98. Within the RL family, RL-SDR-L yields the best PV-only result, with community savings of \texteuro{}734.23, followed by RL-SDR-F (\texteuro{}616.26) and RL-M (\texteuro{}418.49). The SDR-shaped formulations therefore outperform the multiplier-based parameterization considered, but they do not surpass the rule-based benchmarks in the base PV-only case.

This difference arises from the settlement design rather than from the quality of learning. All three benchmarks are budget-balanced. Bill-sharing redistributes the community benefit ex post, while MMR and SDR settle every internal transaction at a single price, such that every payment made by a buyer is received by a seller and the entire grid-price spread remains within the community. Their community savings are therefore determined by the traded volume and the grid price corridor alone, and the three values accordingly coincide at \texteuro{}829.98 even though they price and allocate the exchange differently. The RL modes instead quote separate buy and sell prices, and the resulting spread is, for a given traded volume, a net outflow from the participants. Budget-balanced settlement therefore sets an upper bound on the community savings that any mechanism retaining an internal spread can attain on the same physical exchange. The same pattern appears in the reported costs and revenues, where MMR attains \texteuro{}95.75 higher community savings than RL-SDR-L, with lower user cost (\texteuro{}6,043.68 vs \texteuro{}6,095.92) and higher prosumer revenue (\texteuro{}2,125.85 vs \texteuro{}2,082.34). The same ordering holds operationally in Table~\ref{tab:results_operational}, where traded energy is the cleared internal volume, counted once per transaction, and the self-sufficiency index is evaluated at community level from aggregate grid imports and aggregate household consumption. In PV-only, the rule-based mechanisms clear the full internally matchable volume (7,266~kWh) and reach the highest self-sufficiency (0.304), whereas the RL policies clear slightly less (RL-SDR, 7,131~kWh) or substantially less (RL-M, 4,069~kWh). The shortfall arises in the intervals where the settlement condition fails. Under RL-SDR, the ex-ante buy price reaches the grid import price in the interval that follows an hour without local surplus, which prevents internal settlement in that interval, and a small additional volume is not settled internally in the intervals where the grid price corridor itself is inverted. Under RL-M, the volume not settled internally arises where the paired multipliers invert the internal spread, and it combines two distinct effects. At these tariffs, 1,768~kWh of the matchable volume falls in intervals where no pair in the action set yields a valid spread, and this part of the shortfall hence follows from the action set rather than from the learned policy. The remaining 1,429~kWh is not settled internally in intervals where at least one feasible pair existed and the learned policy nevertheless selected an inverted one.

When the community moves from PV-only to PV-BES, savings increase from \texteuro{}418.49 to \texteuro{}606.94 for RL-M, from \texteuro{}616.26 to \texteuro{}860.09 for RL-SDR-F, and from \texteuro{}734.23 to \texteuro{}978.52 for RL-SDR-L. Storage therefore improves the economic performance of the learning-based policies, especially the SDR-shaped variants. Because $C^{\mathrm{ref}}$ excludes the battery, these figures combine two effects. With batteries operating but no internal trading, the community cost falls to \texteuro{}4,642.05. Accordingly, \texteuro{}105.76 of the reported PV-BES savings comes from storage alone, before accounting for the charging energy committed upstream, and would accrue under any settlement mechanism. The internal settlement produces the remaining \texteuro{}872.76 under RL-SDR-L, \texteuro{}138.53 more than the same policy achieves in the PV-only case.

Fig.~\ref{fig:participation_sdr_l_pvbes} provides an operational view of the best-performing RL policy at household level, with negative values denoting net energy sold to the local market. On the day shown, selected for its high battery activity, prosumer households act predominantly as net sellers during the solar production window, while consumer households absorb that local surplus as net buyers. Both panels use the same household demand profiles and PV portfolio, and the two trained policies share the fixed buy-price sensitivity $\beta$, and the difference between the panels thus reflects battery operation alone. In the PV-BES panel, trading therefore extends beyond the solar window as the scheduled battery discharge is exchanged in the evening hours, consistent with the higher traded energy and self-sufficiency reached by RL-SDR-L in that configuration.

In both PV-only and PV-BES, RL-SDR-F and RL-SDR-L attain the same traded energy and self-sufficiency index according to Table~\ref{tab:results_operational}. The advantage of RL-SDR-L therefore lies in a more effective economic settlement of the same internal exchange, not in a larger traded volume. That settlement advantage is measurable, as the annual internal spread falls from \texteuro{}210.46 under RL-SDR-F to \texteuro{}92.49 under RL-SDR-L in the PV-only case, a reduction of \texteuro{}117.97 that matches the community-savings difference exactly. The learned policy is state-dependent rather than constant, selecting a mean $\alpha$ of 0.26 in surplus intervals and 0.48 in deficit intervals. The same settlement advantage holds in PV-BES, where RL-SDR-L exceeds RL-SDR-F by \texteuro{}118.43 on the same traded volume. The rule-based benchmarks are not re-evaluated under storage in this study, and the PV-BES results consequently compare the learning-based modes only with one another and are not directly comparable with the PV-only rule-based values. Because the battery trajectory is fixed independently of the settlement mechanism, the budget-balance bound above indicates that the rule-based benchmarks would retain their advantage under storage.

\begin{figure}[t]
    \centering
    \includegraphics[width=\textwidth]{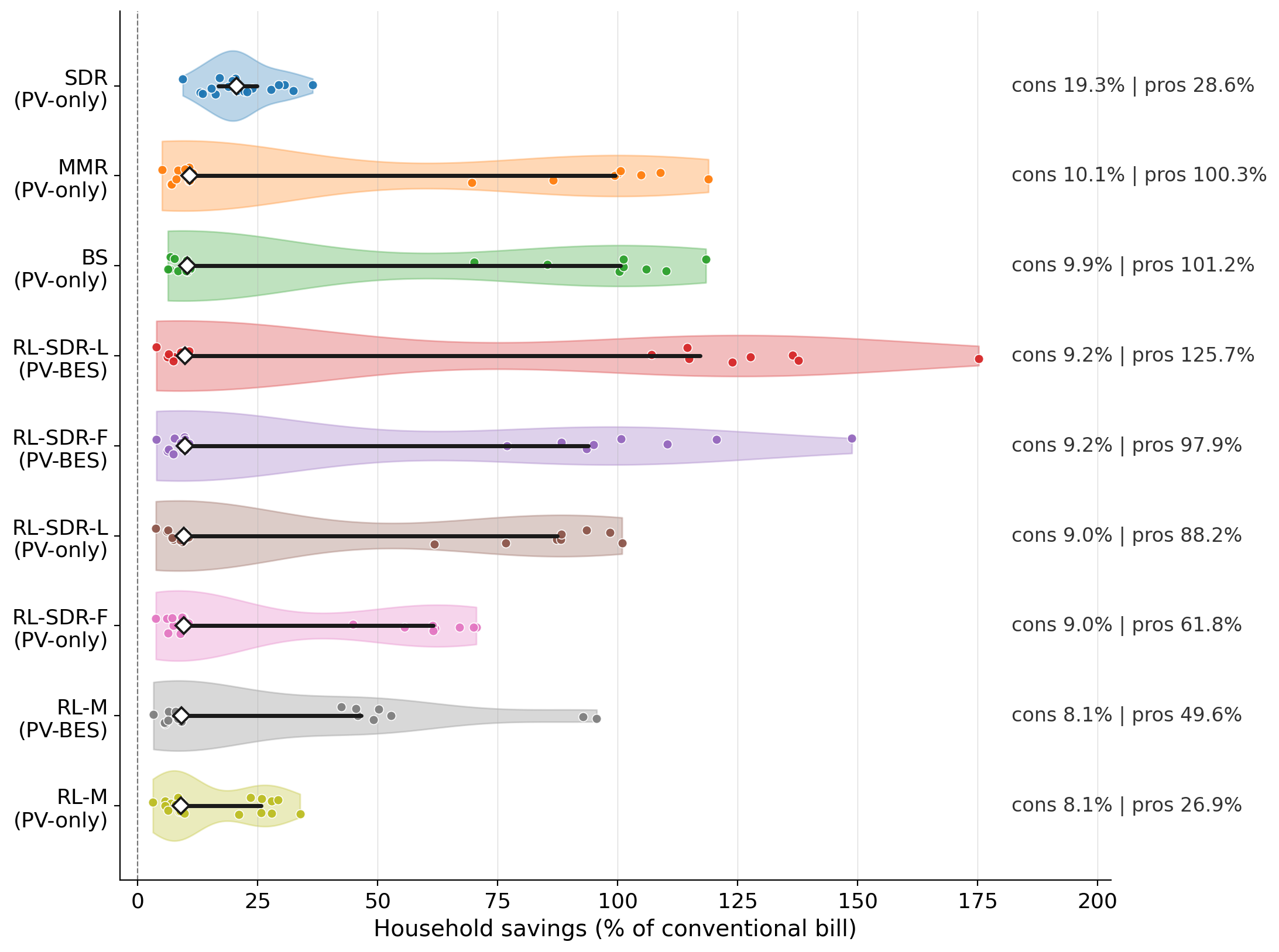}
    \caption{Distribution of household savings by settlement mechanism.}
    \label{fig:savings_distribution}
\end{figure}

\subsection{Household-Level Distribution of Benefits}

Household-level outcomes show that aggregate community gains are not distributed uniformly across participants, as illustrated in Fig.~\ref{fig:savings_distribution}. Each distribution collects the savings of the 20 households under one mechanism, expressed relative to each household's conventional net bill. The diamond marks the pooled median, the horizontal bar marks the interquartile range, and the annotation reports the median saving of the consumer group and of the prosumer group. For prosumers the conventional bill is reduced by export revenue, which allows savings to exceed 100\%, in which case the annual net bill turns into a small net income. The rule-based mechanisms appear for the PV-only configuration, while the RL modes appear for both configurations. In the PV-only framework, the three benchmarks deliver identical aggregate community savings, yet they distribute those savings differently across households. SDR distributes the savings most evenly between the two groups, with consumer and prosumer medians of 19.3\% and 28.6\%, whereas BS and MMR give consumer medians near 10\% and prosumer medians above 100\% on the same aggregate community savings. The pooled median mainly reflects the twelve consumer households, while the extended upper tails correspond to prosumers, whose percentages rest on small net-bill denominators.

A similar pattern appears within the RL family. In PV-only, RL-SDR-L reaches \texteuro{}734.23, above RL-SDR-F (\texteuro{}616.26) and RL-M (\texteuro{}418.49). After storage is added, these values rise to \texteuro{}978.52, \texteuro{}860.09, and \texteuro{}606.94, respectively, yet the gain is not distributed evenly across households. In PV-BES, the RL-SDR variants develop a more extended upper tail, which indicates stronger gains for only part of the community rather than for all households in the same proportion. This household-level behavior matches the community-level results, where rule-based benchmarks remain competitive in PV-only, SDR-shaped RL pricing outperforms the multiplier-based RL mode, and RL-SDR-L remains the best-performing learning-based mode under battery storage.

\section{Conclusions}\label{sec:conclusions}

This paper compared rule-based and RL-based pricing mechanisms for P2P electricity trading in residential PV communities. In the base PV-only configuration, the rule-based benchmarks outperformed the best RL policy. With battery storage, the best RL policy increased community savings from \texteuro{}734.23 to \texteuro{}978.52 before accounting for the energy used to charge the batteries. The rule-based benchmarks were evaluated only in the PV-only configuration. The budget-balance bound established in Sect.~\ref{sec:results} indicates that they would retain their advantage under storage, although this comparison was not carried out in this study.

Within the RL family and in both configurations, the SDR-shaped policies outperformed the multiplier-based parameterization considered. The advantage of RL-SDR-L over RL-SDR-F is not explained by a larger traded volume or a higher self-sufficiency index, but by a more effective economic settlement of the same internal exchange. These findings support adaptive SDR-based pricing for local electricity markets, particularly when PV communities include storage. The settlement does not represent network charges, levies, or taxes. The reported savings are therefore upper bounds in this respect.

Future work may extend the analysis to seasonal operating conditions, evaluate the rule-based benchmarks under storage, and compare rule-based and learning-based pricing under network-constrained and inter-community P2P trading.

\section*{Acknowledgements}
The research was conducted as part of the CoEnerBuild project, funded by the Polish National Center for Research and Development and the European Union under the CETPartnership program (project code Cetp-FP-2023-00014). It was also partially conducted as part of the statutory research activity of the Mineral and Energy Economy Research Institute of the Polish Academy of Sciences.

\section*{Declaration of competing interest}
The authors have no competing interests to declare that are relevant to the content of this article.

\bibliographystyle{elsarticle-num}
\bibliography{references}

\end{document}